\documentclass{article}
\usepackage{spconf,amsmath,graphicx}
\usepackage{multirow}
\usepackage{amsfonts}

\title{M-MELD: A Multilingual Multi-Party Dataset for Emotion Recognition in Conversations}

\name{\begin{tabular}{c} Sreyan Ghosh$^{1\star}$,\qquad S Ramaneswaran$^{2\star}$,\qquad Utkarsh Tyagi$^{3\star}$,\qquad Harshvardhan Srivastava$^{4}$, \\ Samden Lepcha$^{3}$,\qquad S Sakshi$^{5}$,\qquad Dinesh Manocha$^{1}$\thanks{\hspace*{-1mm}$^{\star}$These authors contributed equally to this work}\end{tabular}}

\address{$^1$University of Maryland, College Park, USA, $^2$Nvidia, Bangalore, India
\\ $^3$MIDAS Labs, IIIT Delhi, India, $^4$IIT Delhi, India $^5$UMass Amherst, USA}

\begin{document}

\maketitle
\begin{abstract}
Expression of emotions is a crucial part of daily human communication. Emotion recognition in conversations (ERC) is an emerging field of study, where the primary task is to identify the emotion behind each utterance in a conversation. Though a lot of work has been done on ERC in the past, these works only focus on ERC in the English language, thereby ignoring any other languages. In this paper, we present Multilingual MELD (M-MELD), where we extend the Multimodal EmotionLines Dataset (MELD) \cite{poria2018meld} to 4 other languages beyond English, namely Greek, Polish, French, and Spanish. Beyond just establishing strong baselines for all of these 4 languages, we also propose a novel architecture,  DiscLSTM, that uses both sequential and conversational discourse context in a conversational dialogue for ERC. Our proposed approach is computationally efficient, can transfer across languages using just a cross-lingual encoder, and achieves better performance than most uni-modal text approaches in the literature on both MELD and M-MELD. We make our data and code publicly on GitHub \footnote{https://github.com/Sreyan88/M-MELD}.
\end{abstract}

\section{Introduction}
\label{sec:intro}

In the past few years, ERC has become increasingly popular as an emerging research topic in Natural Language Processing (NLP), with potential applications in areas of empathetic dialog systems \cite{majumder2020mime}, improved human-computer interaction \cite{cowie2001emotion}, and social media opinion mining \cite{chatterjee2019semeval}. Researchers have proposed several methodologies \cite{ghosal2019dialoguegcn,majumder2019dialoguernn,ghosal2020cosmic} to solve this task and have primarily evaluated their systems on 4 popular benchmark datasets, namely MELD \cite{poria-etal-2019-meld}, IEMOCAP \cite{busso2008iemocap}, DailyDialog \cite{li-etal-2017-dailydialog}, and EmoryNLP \cite{zahiri2018emotion}. ERC, particularly emotion recognition in multiparty conversation (ERMC), is more challenging than tasks like sentiment classification due to the presence of emotional dynamics in conversations \cite{poria-etal-2019-meld}.

Though a lot of datasets and systems have been proposed for ERC, these datasets are in English, thereby ignoring any other languages. Different languages function in different ways. Building uni-modal text ERC systems for languages beyond English is especially challenging owing to the fact that recent literature shows how multi-modal cues \cite{hu-etal-2021-mmgcn} and external knowledge \cite{ghosal2020cosmic,li2021past} is important for ERC. Multi-modal systems, though outperform text-only systems, may not be applicable in all real-world scenarios like chatbots, where only textual information is available. For text-only ERC, external cues like commonsense \cite{li2021past} or psychological knowledge \cite{ghosal2020cosmic} depend on external sources of knowledge like knowledge graphs, which restricts their use to ERC models trained on English due to the lack of these sources in any other language.




{\noindent \bf Main Contributions}:  In this paper we attempt to bridge this gap between English and non-English languages for ERC by first proposing a new dataset, M-MELD. M-MELD consists of over 4504 human-annotated text dialogues and 45871 individual utterances in 4 different languages, namely Greek, Polish, French, and Spanish, spanning 7 distinct emotions and balanced equally across 2 high-resource (French and Spanish) and 2 low-resource (Greek and Polish) languages. We describe the annotation procedure in Section \ref{sec:mmeld}, where we also describe why in-context human-translated dialogues are required for training better ERC systems and training over machine translated (MT) dialogues results in worse performance. Second, we also present a new model for ERC namely \textbf{DiscLSTM}, or Discourse-aware LSTM, which incorporates discourse-aware structured graph information into sequential recurrence-based learning. Precisely, we first train a discourse parsing model using cross-lingual embeddings from XLM-RoBERTa \cite{conneau2019unsupervised} on a popular human-annotated dataset annotated for discourse relations between human utterances \cite{asher-etal-2016-discourse}. Next, we infer discourse relations for dialogues in multiple languages and use these cues to incorporate a long-distance conversational background in DiscLSTM. More information on DiscLSTM can be found in Section \ref{sec:disclstm}. To sum up, our main contributions are as follows:

\begin{itemize}
    \item We propose M-MELD, the first dataset for ERC in languages beyond English. M-MELD has over 45871 utterances in 4504 dialogues with human-translated utterances for ERC and is balanced across languages. Our proposed dataset is also useful in multilingual sentiment classification and as a human-annotated parallel corpus for learning MT systems. Additionally, we establish strong baselines for ERC with M-MELD.
    \item We propose DiscLSTM, a simple yet powerful model for ERC. Unlike most state-of-the-art systems in the literature, DiscLSTM can be easily adapted across languages and does not require external knowledge, is more resource friendly, and outperforms our text-based uni-modal baselines taken from literature.
\end{itemize}

\section{Related Work}
\label{sec:related}

One of the first works in this space was DialogRNN
\cite{majumder2019dialoguernn}, which proposed modeling dialog dynamics with stacked RNNs. Following DialogRNN, the same authors proposed DialogueGCN \cite{ghosal2019dialoguegcn}, which treats each dialogue as a graph, with vertices as individual utterances, and edges connecting a vertex with its past and future turns. DAG-ERC \cite{shen-etal-2021-directed} uses a Directed Acyclic Graph, which combines the benefits of graph and recurrence models with its structural properties. Very recently, MMGCN \cite{hu-etal-2021-mmgcn} proposed fusing information from multiple modalities by the use of spectral domain GCN to encode the multimodal contextual information. The work closest to DiscLSTM is \cite{sun2021discourse}, where the authors use discourse relations between utterances to build a conversational graph and show that ER in both multi-party and two-party conversations benefit from conversational discourse structures. Another popular system is DialogXL which uses dialog-aware self-attention. Finally, CoMPM \cite{lee-lee-2022-compm} combines context embedding and pre-trained speaker memory to reflect the dialogue context and EmotionFlow \cite{emotionflow} learns user-specific features to model the spread impact of emotion in a conversation.


\section{Multilingual MELD (M-MELD)}
\label{sec:mmeld}

In this section, we describe in detail the annotation procedure for M-MELD. MELD is a dataset for ERMC. In addition to the text transcript of the utterance, MELD also consists of acoustic and visual cues for each utterance. MELD has more than 1400 dialogues and 13000 utterances from the TV series Friends. Each utterance in dialogue has been labeled one of these seven emotions:- Anger, Disgust, Sadness, Joy, Neutral, Surprise, and Fear. In this paper, and for M-MELD, we are only concerned with the text modality for each utterance.

The most trivial procedure to curate M-MELD would be to use state-of-the-art Machine Translation (MT) models to translate MELD utterances into any desired language. However, MT systems largely rely on the assumption that sentences can be translated in isolation. Indeed, we hypothesize that translation of conversations or dialogues requires translating sentences coherently with respect to conversational flow in order for all aspects of the information exchange, including speaker intent, attitude, and style, to be correctly communicated. Thus, as a first step, we curate human-annotated translations for M-MELD from foreign language subtitles \footnote{https://www.tvsubtitles.net/tvshow-65-1.html}. The next step involves finding the exact utterances in the subtitles that match the utterances in MELD. Though MELD has timestamps for each utterance relative to each episode, it should be noted that these timestamps are prone to changing when in another foreign language, as utterance durations might shorten or lengthen. Thus, we hire 4 professional annotators (1 for each language) to manually match each utterance in MELD dialogues to their actual utterances in the subtitles. To reduce the annotation burden, we  show them English utterances that occur at a maximum of +-10s from the foreign utterance to be annotated, as the original sentence was guaranteed to be found within that window. Statistics for our M-MELD dataset can be found in Table \ref{table:stats_1}.


\begin{table}[t]
\caption{Dataset Statistics for M-MELD}
\vspace{1mm}
\begin{tabular}{|c|ccc|ccc|}
\hline
\multirow{2}{*}{Language} & \multicolumn{3}{c|}{\# Dialogues}                            & \multicolumn{3}{c|}{\# Utterances}                            \\ \cline{2-7} 
                          & \multicolumn{1}{c|}{Train} & \multicolumn{1}{c|}{Dev} & Test & \multicolumn{1}{c|}{Train} & \multicolumn{1}{c|}{Dev}  & Test \\ \hline
French                    & \multicolumn{1}{c|}{633}   & \multicolumn{1}{c|}{97}  & 224  & \multicolumn{1}{c|}{6537}  & \multicolumn{1}{c|}{964}  & 2198 \\ \hline
Greek                     & \multicolumn{1}{c|}{870}   & \multicolumn{1}{c|}{103} & 240  & \multicolumn{1}{c|}{9003}  & \multicolumn{1}{c|}{1062} & 2366 \\ \hline
Spanish                   & \multicolumn{1}{c|}{769}   & \multicolumn{1}{c|}{111} & 268  & \multicolumn{1}{c|}{7890}  & \multicolumn{1}{c|}{1064} & 2546 \\ \hline
Polish                    & \multicolumn{1}{c|}{858}   & \multicolumn{1}{c|}{96}  & 235  & \multicolumn{1}{c|}{8928}  & \multicolumn{1}{c|}{989}  & 2324 \\ \hline
\end{tabular}
\label{table:stats_1}
\end{table}

\section{Discourse-aware LSTM}
\label{sec:disclstm}

\subsection{Problem Formulation}
\label{sec:problem_form}
The general problem of ERC can be formulated as follows. Suppose there are $m$ participants \{$p\textsubscript{1}, p\textsubscript{2}, p\textsubscript{2}, \cdots, p\textsubscript{m}$\} in a conversation or dialogue with $n$ number of utterances \{$e\textsubscript{1}, \cdots, e\textsubscript{i}, \cdots, e\textsubscript{n}$\}, where utterance $e\textsubscript{i}$ is uttered by $p(e\textsubscript{i})$, and $p(.)$ denotes the mapping between an utterance and it's speaker. The primary objective of ERC is to predict the emotion label $y\textsubscript{i}$ for utterance $e\textsubscript{i}$ based on the context of the dialogue to which $e\textsubscript{i}$ belongs. We denote a dialogue as $U\textsubscript{j}$, where utterance $e\textsubscript{i} \in U\textsubscript{j}$ and $D = \{U\textsubscript{1}, \cdots, U\textsubscript{j}, \cdots, U\textsubscript{T}$\}, where dataset $D$ has a total of $T$ dialogues.

\subsection{Utterance-level Feature Extraction}
\label{sec:utter_feature_extraction}
We formulate each conversation $U_j$ as a graph and treat each utterance embedding $u\textsubscript{i} \in U_j$ as a node in the graph. In line with prior methods \cite{ghosal2019dialoguegcn,shen-etal-2021-directed}, we use the XLM-RoBERTa$_{large}$ transformer model, fine-tuned on the task-specific ERC dataset to extract sentence-level features $u\textsubscript{i} \in \mathbb{R}^{1024}$ for each individual utterance $e_i$ in the dialogue. More precisely, similar to \cite{ghosal2019dialoguegcn}, we add a \emph{[CLS]} token at the beginning of each tokenized utterance we feed into our XLM-RoBERTa$_{large}$ model; the output embedding of the \emph{[CLS]} acts as our pooled utterance representation. Conversation $U_j$ can now be denoted by $U_j$ = \{$u\textsubscript{1}, \cdots u\textsubscript{i}, \cdots, u\textsubscript{n}$\} or a set of utterance representations from XLM-RoBERTa$_{large}$.

\subsection{Dialogue Discourse Parsing}
\label{sec:discourse_parsing}
For dialogue discourse parsing, we first train a state-of-the-art model \cite{liu2021improving} on a human-annotated multi-party dialogue corpus STAC \cite{asher-etal-2016-discourse}. We view each utterance as an EDU (Elementary Discourse Unit) and use the discourse relation types defined in STAC. \cite{liu2021improving} employs a transformer backbone. One simple trick that enables us to obtain discourse relations in dialogues across languages is using cross-lingual embeddings from XLM-RoBERTa to train our discourse parsing model. Thus, even though STAC is in English, the model supports a wide range of languages during inference time.

\subsection{Disc-LSTM Architecture}
\label{sec:disc_lstm_arch}

\subsubsection{Conversation Graph Construction}
\label{sec:conve_graph_construc}

With discourse relations obtained from the previous step, we construct a discourse graph $G$ = ($V$,$E$) for each conversation, where $V$ = \{$v$\textsubscript{$1$}, $v$\textsubscript{$2$}, $v$\textsubscript{$3$}, $\cdots$, $v$\textsubscript{$n$}\} are the vertices or nodal representations of the utterances \{$u$\textsubscript{$1$}, $u$\textsubscript{$2$}, $u$\textsubscript{$3$}, $\cdots$, $u$\textsubscript{$n$}\} and $E \in \mathbb{R}^{n \times n}$ is the adjacency matrix denoting edge relation, where $E[i][j]$ = 1 if there is a discourse relation between utterances $i$ and $j$. 

\subsubsection{Temporal Information Flow in Graph Layers}
\label{sec:temp_inform_flow}

To encode discourse relations in a conversation, we use a Graph Attention Network (GAT) \cite{velivckovic2017graph}  with the information flow through layers inspired by \cite{DBLP:journals/corr/abs-2105-12907}. To feed our contextualized RoBERTa-based utterance embedding $u_i$ $\in$ $\mathbb{R}^{1024}$ to our graph network, we first down-project $e_i$ to $g_{i}^{1}$, where a $g_{i}^{1}$ $\in$ $\mathbb{R}^{300}$ via a full-connected layer $f(.)$ as follows: $G^1 = f(U) = \{g_{1}^{1},g_{2}^{1},g_{3}^{1},\cdots,g_{n}^{1}\}$, where $G^1$ is the graph-encoded representation of our utterance output by the first layer in the graph. For each utterance embedding $g$\textsubscript{$i$}, the attention weights between $g$\textsubscript{$i$} and its predecessors are calculated by using $g$\textsubscript{$i$}’s hidden state at the $(l-1)$-th layer and the nodes $j \in \mathcal{N}_{i}$ in the current $(l)$-th. Formally, we find the attention weights of utterance $u$\textsubscript{i}'s hidden value with the above-mentioned nodes in the following manner using a GAT layer:

\begin{equation}
    \alpha_{ij}^{l} = \text{softmax}_{j \in \mathcal{N}_{i}}(W_{\alpha}^{l}[g^{l}_{j}||g^{l-1}_i])
\end{equation}
where $W_{\alpha}^{l}$ are the learnable parameters and $||$ represents a concatenation operation. We finally gather or accumulate the information using the weights calculated above and get the subsequent layer information by: $g^{l}_{i} = \sum_{j \in \mathcal{N}\textsubscript{i}} \alpha_{ij}g_{j}^{l} + g_{i}^{l-1}$.
We use the final graph layer embedding after multiple information propagation steps in the graph network and obtain  $G^l=\{g_{1}^{l},g_{2}^{l},g_{3}^{l},\cdots,g_{n}^{l}\}$ and the contextualized embeddings from the RoBERTa for our next step.


\begin{figure}[t]
  \centering
\includegraphics[width=8cm,height=11cm,trim={0cm 0.5cm 0cm 0cm}]{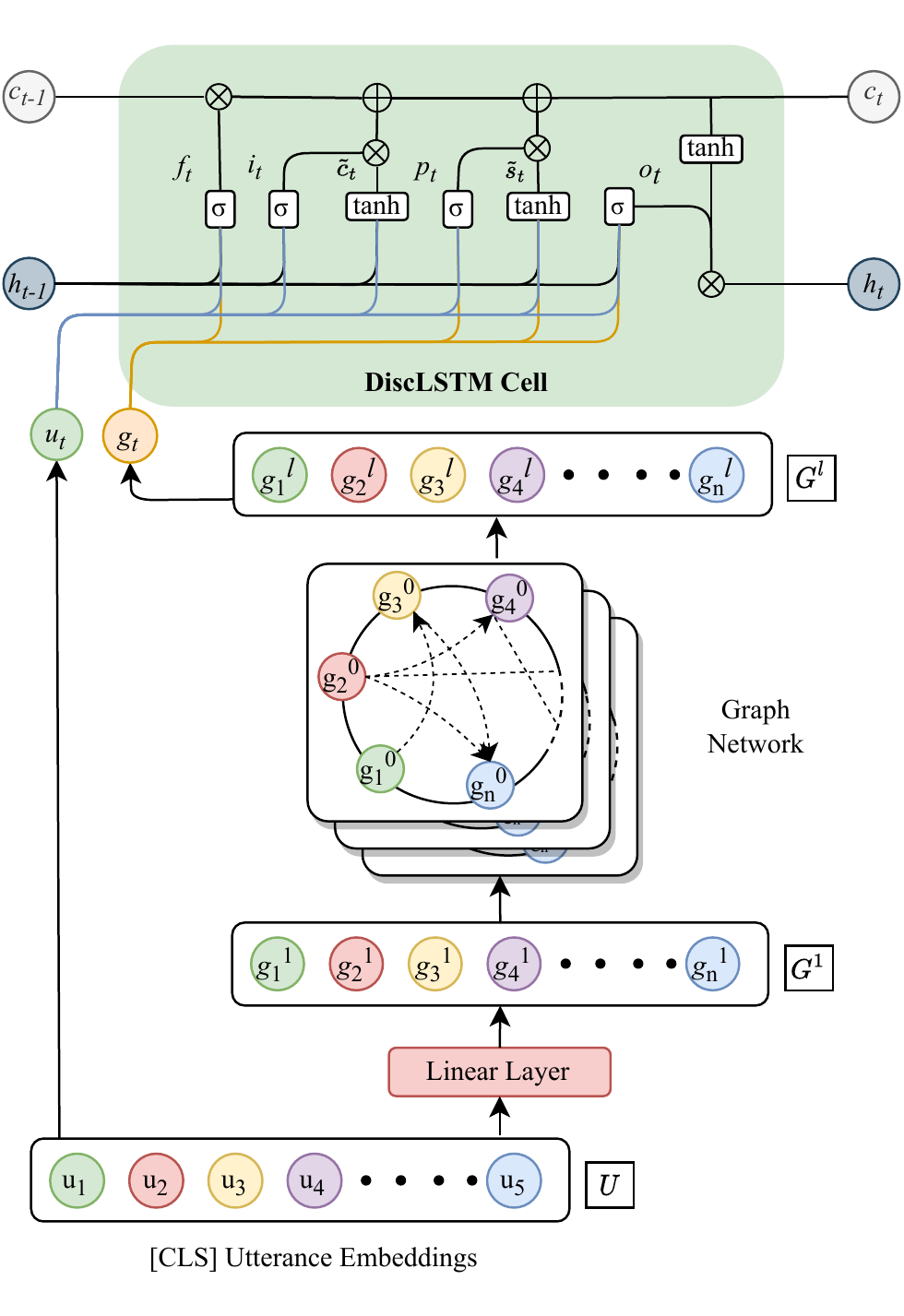}
  \caption{Illustration of our proposed DiscLSTM. The utterance embedding $u$ is passed through a Graph Encoder $G$ post which it is passed to the DiscLSTM cell together with the output of the graph encoder $g^l$.}\label{fig:figure_2}
\end{figure}

\subsubsection{Bi-Directional DiscLSTM Cell}
\label{sec:bidirectional_lstm_cell}

In order to better integrate both the sequential and discourse context and to dynamically learn the relative importance of each graph-encoded utterance representation for modeling a conversation, we propose \textbf{DiscLSTM}. DiscLSTM builds on the basic LSTM cell and takes as inputs previous cell state $c_{t-1}$, previous hidden state $h_{t-1}$, current cell input $u_{t}$, and an additional graph-encoded utterance representation $g_{t}$. The cell outputs the current cell state $c_{t}$ and the current hidden state $h_{t}$. The cell representation can be seen in Fig. \ref{fig:figure_2}. The following represents the propagation of information inside the DiscLSTM cell:\\

$
\begin{aligned}
\mathbf{f}_{t} &=\sigma\left(W_{(f)} \mathbf{u}_{t}+U_{(f)} \mathbf{h}_{t-1}+Q_{(f)} \mathbf{g}_{t}+\mathbf{b}_{(f)})\right.\\
\mathbf{o}_{t} &=\sigma\left(W_{(o)} \mathbf{u}_{t}+U_{(o)}
\mathbf{h}_{t-1}+Q_{(o)} \mathbf{g}_{t}+\mathbf{b}_{(o)}\right) \\
\end{aligned}
$

$
\begin{aligned}
\mathbf{i}_{t} &=\sigma\left(W_{(i)} \mathbf{u}_{t}+U_{(i)} \mathbf{h}_{t-1}+\mathbf{b}_{(i)}\right) \\
\mathbf{p}_{t} &=\sigma\left(W_{(p)} \mathbf{u}_{t}+Q_{(p)} \mathbf{g}_{t}+\mathbf{b}_{(p)}\right) \\
\tilde{\mathbf{c}}_{t} &=\tanh \left(W_{(u)} \mathbf{u}_{t}+U_{(u)} \mathbf{h}_{t-1}+\mathbf{b}_{(u)}\right) \\
\tilde{\mathbf{s}}_{t} &=\tanh \left(W_{(n)} \mathbf{u}_{t}+Q_{(n)} \mathbf{g}_{t}+\mathbf{b}_{(n)}\right) \\
\mathbf{c}_{t} &=\mathbf{f}_{t} \odot \mathbf{c}_{t-1}+\mathbf{i}_{t} \odot \tilde{\mathbf{c}}_{t}+\mathbf{p}_{t} \odot \tilde{\mathbf{s}}_{t} \\
\mathbf{h}_{t} &=\mathbf{o}_{t} \odot \tanh \left(\mathbf{c}_{t}\right)
\end{aligned}
$
\\

where $u_t$ ($u_i$ w.r.t. utterance in a dialogue) is the XLM-RoBERTa\textsubscript{large} utterance representation and $g_t$ ($g_{i}^{l}$ w.r.t. utterance in a dialogue) is the graph-encoded utterance representation for the $t^{th}$ time-step in the sequential processing by the bi-directional DiscLSTM cell. The forward and backward DiscLSTM enable the model to integrate both the sequential and structured-discourse information from both directions in the sequence. Finally, we concatenate the hidden
state $\overrightarrow{h_{if}}$ and hidden state $\overleftarrow{h_{ib}}$ from the forward and backward states respectively to get the final hidden state representation of $i$th utterance $h_{i} = [\overrightarrow{h_{if}};\overleftarrow{h_{ib}}]$. This final hidden state representation $\textbf{H}=\{h_{1},h_{2},h_{3},\cdots,h_{n}\}$ is then fed to a fully-connected layer which outputs a vector representation $p_i$ $\in$ $\mathbb{R}^{d}$ for each utterance $e_i$ where $d$ equals the number of emotion classes in the ERC dataset.

\section{Experiments}
\label{sec:exp}

{\textbf{Baselines:}} For evaluating systems from prior-art across languages in M-MELD and also to compare DiscLSTM, we choose systems that do not require an external source of knowledge, e.g., knowledge graphs, and can be easily re-implemented across languages. To the best of our knowledge, external sources of knowledge commonly used in recent literature attributing to their success \cite{ghosal2020cosmic,li2021past} are not readily available in foreign languages beyond English. Adding to this, systems like DialogXL depend on large-scale language-specific pre-trained models and their foreign language counterparts are neither available nor resource-friendly to pre-train from scratch. On the other hand, systems like MMGCN \cite{hu-etal-2021-mmgcn} leverage multiple modalities, which is not always available in a real-world setting.

Our first baseline is an XLM-RoBERTa transformer trained on a sequence classification task for ER. We adopted DialogueRNN, DialogueGCN and DAG-ERC described in Section \ref{sec:related} to work with utterance-level features extracted using the methodology in Section \ref{sec:utter_feature_extraction}. Additionally, we fine-tune EmotionFlow and CoMPM end-to-end with XLM-RoBERTa backbone for each language. To prove the need for human-translated annotations, we also evaluate performance of all our baselines on synthetic MT data.

\vspace{1.5mm}
{\noindent \textbf{Experimental Setup:}} We use the pre-trained XLM-RoBERTa from the Huggingface library. For training and evaluation of all our systems, including baselines and DiscLSTM, we use a batch size of 16 and train our networks for 50 epochs using Adam optimizer with a learning rate of $1e^{-5}$. For training XLM-RoBERTa$_{large}$ we find an optimal learning rate of $1e^{-4}$. Optimal hyperparameters for all other baselines were obtained from their respective papers. 


\section{Results}

Following much of prior-art, we evaluate the performance of all our baselines and our proposed DiscLSTM on the \emph{weighted $F_{1}$} score. Results are presented in Table \ref{tab:results}. As we clearly see, DiscLSTM outperforms most of our baselines on all 4 languages in M-MELD and is close to DialogueGCN and DAS-ERC in French and Spanish. However, one must note that unlike DialogueGCN and DAG-ERC, DiscLSTM does not use future utterance information for ERC. In addition to M-MELD, we also repeat our baselines for the original MELD and the results are 62.10 / 63.16 / 63.5 / 52.81 / 50.84 / \textbf{63.75} respectively in same order as in Table \ref{tab:results}. DiscLSTM is also more efficient than all our graph-based baselines as it uses much lesser edges on average than our other baselines which extend an edge from each utterance to all future and past utterances. Additionally, we notice an drop of 2.1\% average across methods when trained on MT data. Detailed results can be found on 
GitHub.


\begin{table}[!h]
\caption{Results on the various languages in M-MELD}
\label{tab:results}
\vspace{1mm}
\begin{tabular}{l|c|c|c|c}
\hline
\textbf{Model} &\textbf{French} &\textbf{Spanish} &\textbf{Greek} &\textbf{Polish}\\
\hline
XLM-RoBERTa &49.14 &52.30 &52.70& 31.0 \\
DialogueRNN \cite{majumder2019dialoguernn} &50.19 &52.50 &53.10&31.41 \\
DialogueGCN \cite{ghosal2019dialoguegcn} &\textbf{51.20} &52.81 &53.46 &31.90 \\
CoMPM \cite{lee-lee-2022-compm} &35.41 &49.00 &42.76 & 37.03 \\
EmotionFlow \cite{emotionflow} &35.69 & 41.65 &43.13 & 42.20 \\
DAG-ERC \cite{shen-etal-2021-directed} &49.10 & 52.9 &53.40 & 42.23\\
\textbf{DiscLSTM} (ours) & 49.43 & \textbf{53.24} & \textbf{53.46} & \textbf{43.21} \\
\hline
\end{tabular}
\end{table}

\section{Conclusion}
\label{sec:conclusion}

In this paper, we present M-MELD, the first publicly available multilingual dataset for ERC in languages beyond English, namely French, Spanish, Greek, and Polish. We establish strong baselines from literature for all 4 languages in M-MELD and discuss how most modern systems achieving state-of-the-art in MELD get restricted to their usage just in English. Our dataset opens new challenges to the research community in designing ERC systems that can be easily transferred across languages. Additionally, we also propose a new and efficient system for ERC, DiscLSTM, which outperforms all our unimodal text baselines across MELD and M-MELD. A limitation of DiscLSTM is the two-step approach where error does not propagate across stages. As part of future work we would devise better end-to-end language-independent ERC systems. 



\vfill\pagebreak

\bibliographystyle{IEEEbib}
\bibliography{strings,refs}

\end{document}